# Learn Task First or Learn Human Partner First: A Hierarchical Task Decomposition Method for Human-Robot Cooperation

Lingfeng Tao*, Michael Bowman *, Jiucai Zhang^, and Xiaoli Zhang*, *Member, IEEE*

*Abstract*—Applying Deep Reinforcement Learning (DRL) to Human-Robot Cooperation (HRC) in dynamic control problems is promising yet challenging as the robot needs to learn the dynamics of the controlled system and dynamics of the human partner. In existing research, the robot powered by DRL adopts coupled observation of the environment and the human partner to learn both dynamics simultaneously. However, such a learning strategy is limited in terms of learning efficiency and team performance. This work proposes a novel task decomposition method with a hierarchical reward mechanism that enables the robot to learn the hierarchical dynamic control task separately from learning the human partner's behavior. The method is validated with a hierarchical control task in a simulated environment with human subject experiments. Our method also provides insight into the design of the learning strategy for HRC. The results show that the robot should learn the task first to achieve higher team performance and learn the human first to achieve higher learning efficiency.

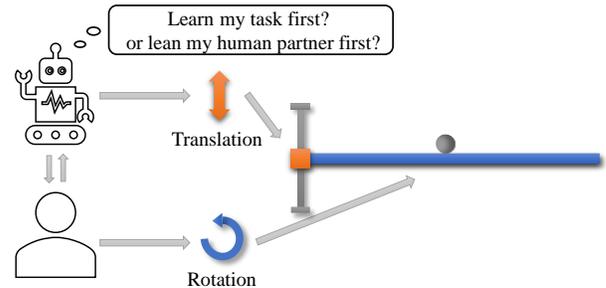

Fig. 1. In this HRC case, the robot and the human cooperatively complete a hierarchical dynamic control task, where the low-level subtasks are to keep the orange slider in the middle of the vertical rod and the blue pendulum in a horizontal position, and the high-level subtask is to keep the gray ball in the middle of the pendulum. The question is whether the robot should learn to accomplish the task first or learn to cooperate with the human first.

## I. INTRODUCTION

HRC has been studied in the past decades in applications such as robot-assisted manufacturing [1], teleoperation [2], life assistance [3], and augments the human partner to complete the task together [4]. The rapid development of DRL [5] has shown that it is feasible to solve complex robot control problems in recent years. DRL problems are modeled as a Markov decision process (MDP) [6], where the robot interacts with the environment and receives observations and rewards. Then the agent takes actions based on the feedback information to maximize its performance in the task. Such learning mechanisms mimic the human's behaviors when dealing with an unfamiliar task, which makes DRL naturally suitable to solve HRC.

A potential problem with applying DRL in HRC is that the robot simultaneously learns how to cooperate with the human partner and complete the task [7]. Recent approaches have considered humans as part of the environment [8]. The robot can observe environmental changes and human behaviors. Such scenarios fit simple cooperation tasks like table carrying tasks [9], object manipulation tasks [10], or block stacking tasks [11]. However, in the real world, cooperation usually associates with dynamic environments and complex tasks, including several subtasks, where DRL may not be able to stably and sufficiently learn a cooperative policy [12][13]. The reasons are two-fold. First, the robot observation comprises of end effects caused by both the human and the robot. Human action can hinder the robot from extracting knowledge to build a correct relationship between its action and the corresponding environmental changes. Second, the tedious exploration process of the DRL algorithm may challenge the human partner's patience and distract concentration. As a result, the robot may frustrate the human and lose the trust of the person [14]. The team may take a long time to reach an equilibrium with lower task performance or even fail to complete it.

In this work, we hypothesize that better HRC can be achieved with DRL when the robot learns the task and human partner separately, but **should the robot learn the task first or learn to cooperate with the human partner first (shown in Fig. 1)?** We propose a novel task decomposition method that decomposes the task based on the task priority level and the action executor level with a hierarchical reward mechanism to answer this question. We aim to study the fundamental nature of HRC and develop learning strategies to improve team performance and learning efficiency. The contributions are two-fold:

1) Developed a novel task decomposition method, with a hierarchical rewards mechanism, that allows a robot to use DRL to learn the HRC task with multiple prioritized subtasks and cooperate with the human partner. The proposed method improves learning efficiency and learning outcomes, which results in better cooperation.

2) Understanding the influence of human involvement and tasks with a general asymmetric hierarchical structure in HRC helps develop novel performance evaluation metrics for learning strategy selection.

## II. RELATED WORK

The DRL application to HRC has been studied in the last decade because of its potential ability in complex control tasks [15]. Work to use DRL in control tasks under an HRC setup has improved performance compared with traditional control methods. DRL has enabled the robot to learn and interact with

*L. Tao, M. Bowman, and X. Zhang are with Colorado School of Mines, Intelligent Robotics and Systems Lab, 1500 Illinois St, Golden, CO 80401 USA (e-mail: tao@mines.edu, mibowman@mines.edu, xlzhang@mines.edu).

^J. Zhang is with the GAC R&D Center Silicon Valley, Sunnyvale, CA 94085 USA (e-mail: zhangjiucai@gmail.com).

humans through a trial-and-error method in more straightforward human-robot interaction tasks such as shaking hands and guiding directions [16]. The robot has been equipped with attention-based DRL to interact with many people in a navigation task [17]. The researchers have implemented DRL to handle more complicated decision-making tasks and communication tasks [18][19]. An awareness-based RL algorithm was proposed in [20] to adaptively switch the robot's cooperation level from autonomous to semi-autonomous. In [21], the robot uses a model-based DRL variable impedance controller to assist human partners in a cooperative lifting task. In [22], the DRL method is used for assisted lunar lander game control. The robot acts as a filter and optimizer of the human's control command rather than individually interacts with the environment as in normal HRC. These methods enable the robot to learn to accomplish the task together with the human or to augment human performance. In [23], the authors proposed a probability-based sensorimotor DRL algorithm and used a similar dynamic experiment for validation. In current approaches, the robot and the human are trained together to learn the task and cooperate at the same time. The training strategies on how to improve the learning efficiency and learning outcome for HRC still lack attention. Our work concentrates on using DRL to learn the fundamental nature of HRC in hierarchical dynamic tasks and develop a new perspective of the HRC formalization. The outcome can be used to improve training efficiency and task performance in HRC.

## III. METHODOLOGY

This section explains the development of our approach. Part A introduces the formulation of the HRC problem. Part B introduces the development of our methods. Compared to the methods reviewed in Related Work, the proposed method advances the reward function's design in the HRC task by developing a hierarchical reward mechanism based on the task decomposition tree. It enables the robot to understand the priority relationship of the subtasks during the cooperation.

### A. Problem Formulation

We model the HRC task as an RL problem that follows the MDP. The MDP is defined as a tuple $\{S, A, R, \gamma\}$, where $S$ is the state of the environment, $A$ is the set of robot actions. $r=R(s_{t+1}|s_t, a_t)$ is the reward received after the transition from state $s_t \in S$ to state $s_{t+1} \in S$. $\gamma$ is a discount factor. A policy $\pi(s, \theta)$ specifies the action for state $s$. $\theta$ is the policy network parameters. A Proximal Policy Optimization (PPO) algorithm [24] is adopted to find $\theta$.

### B. Hierarchical Task Decomposition and Reward Mechanism

For a dynamic control task with a hierarchical structure, we first decompose it to different hierarchy levels; each level contains multiple subtasks. The decomposition is tree-based for more straightforward representation and visualization. There are two main components in the decomposition tree. The first one is the subtask, which is denoted as $T_i^j$, where $j=1, 2, ..., J$, which is the index of the logical hierarchy layers. The priority level follows a descending manner as $j$ increases (i.e., the objective with priority 1 has the highest level). $i$ is the subtask index at that level. The second one is action, which is denoted as $A_k^{j'}$, where $j'$ is the index of the level, and $k$ is the index of the action at that level. In HRC, the action can either be executed by the robot or the human. We define the decomposition rules as: 1) A higher-level subtask may have one or multiple lower-level subtask branches. 2) The decomposition tree's root ends are always an action, which means all decomposed subtasks can trace back to actions. According to the decomposition rules, we define two basic components in the decomposition tree: task-action connection and task-task connection, shown in Fig. 2. The decomposition rules help to generate a clear scope of the priorities of subtasks and their relationships.

In HRC, each team member will take control of different actions that correspondingly contribute to the subtasks following the decomposition tree. The human operator can be directly informed of the subtasks and his/her responsibility during the cooperation. Then the human can process the task information and start to cooperate and refine his/her performance. For the robot, the reward function $R$ is essential to guide the learning process. We propose a hierarchical reward mechanism based on the hierarchical task decomposition tree to efficiently guide the robot to learn an optimal policy. For a subtask that is contributed by the robot $T_i^j$, a reward function component $f_i$ is defined as:

$$f_i^j(s \lor a \lor t) \quad (1)$$

$f_i^j$ can be designed as a function of state $s$, action $a$ or time $t$. The overall reward function is cumulated level by level following the decomposition tree:

$$R = \sum_{1}^{j} \sum_{1}^{i} f_i^j \quad (2)$$

In practice, the hierarchical reward mechanism makes it flexible to design the learning strategy for the robot. It should be noted that the human and robot can share the same higher-level tasks, but they do not need to know each other's low-level tasks. With the hierarchical reward mechanism, we design two learning strategies that both have two training steps:

1) *Responsibility-guided cooperative learning.* During the first training step, the robot learns all subtasks that its action can contribute. The human joins at the second training step, cooperating with the robot to complete the overarching task. This learning strategy helps the robot to better understand the environment dynamics during the first stage without the human factor. Then the robot can focus on learning the influence of human involvement and updating its policy to cooperate.

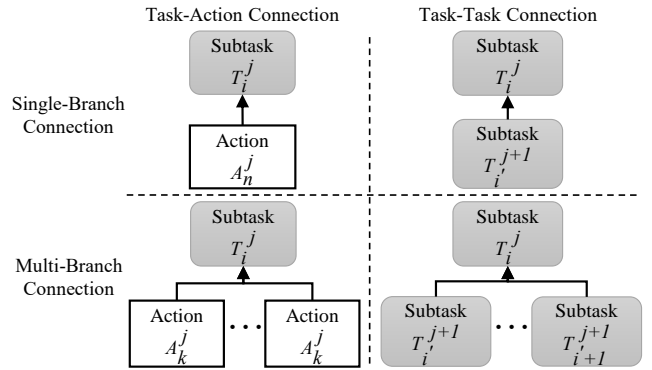

Fig. 2. A hierarchical task that has $j$ levels, each with $i$ subtasks. The level represents the priority of the subtasks.

2) *Level-guided cooperative learning*. At the first training step, the robot and the human learn together from the lowest level subtask. The team starts to learn higher-level subtasks at the next training step if all lower-level subtasks are satisfied. In this learning strategy, the robot can continually learn how to work with the human partner while learning the task. The cooperation starts with relatively simpler and less relevant subtasks, which may help the robot understand human behaviors and adjust its cooperation policy.

Overall, responsibility-guided learning allows the robot to learn the task first, while level-guided learning encourages the robot to learn the human partner first.

## IV. EXPERIMENT

We formulate the HRC problem in an asymmetric hierarchical dynamic task using a slider-pendulum-ball simulator (Fig. 3). The pendulum is attached to a slider with two degrees of freedom: rotation along the *z*-axis and translation along the *x*-axis. A ball rests on the pendulum and can roll along with it. The two available actions are to apply torque to rotate the pendulum and apply force to move the slider up and down. A two-level hierarchical task is designed. There are two subtasks in level 2: $T_1^2$ is to keep the slider in the middle of the rod and $T_2^2$ is to keep the pendulum in the horizontal direction. There is one subtask in level 1: $T_1^1$ is to keep the ball in the middle of the pendulum. Both actions contribute to each subtask in level 2. Both tasks in level 2 contribute to the ball position task in level 1. In this task, each action contributes more to the subtask that is directly contacted (i.e., translation directly controls the slider position but also affects the pendulum's rotational position, vice versa). The pendulum position contributes more than the slider position toward the high-level ball position subtask because the ball is more sensitive to the pendulum's rotational position. The subtasks are intentionally designed to be asymmetric to comprehensively evaluate the developed algorithm and study the HRC with more possible scenarios. The reward component for each subtask for the robot is derived from the Gaussian distribution:

$$f_1^2 = \alpha_1^2 \frac{1}{\sigma_1^2 \sqrt{2\pi}} e^{-\frac{1}{2}\left(\frac{S_x - \mu_1^2}{\sigma_1^2}\right)^2} \quad (3)$$

$$f_2^2 = \alpha_2^2 \frac{1}{\sigma_2^2 \sqrt{2\pi}} e^{-\frac{1}{2}\left(\frac{P_z - \mu_2^2}{\sigma_2^2}\right)^2} \quad (4)$$

$$f_1^1 = \alpha_1^1 \frac{1}{\sigma_1^1 \sqrt{2\pi}} e^{-\frac{1}{2}\left(\frac{B_y - \mu_1^1}{\sigma_1^1}\right)^2} \quad (5)$$

where $\alpha, \sigma, \mu$ are the tunable parameters to shape the reward function. In the experiment, all $\alpha$ were set to 10, all $\sigma$ were set to 1, and all $\mu$ were set to 0. $P_z$ is the rotation angle of the pendulum with east direction as 0 degree, and $S_x$ is the slider's vertical position with the midpoint of the rod as the point 0, $B_y$ is the ball position with the midpoint of the pendulum as the point 0. The length of the pendulum is 0.5m, the length of the rod is 0.2m.

The experiment was implemented in a simulated environment with real human subjects. Constraints were set up on both ends of the pendulum to prevent the ball from falling off and causing unnecessary restarts to the training. The human can visually observe the environment on a monitor and execute the action by controlling a joystick (Fig. 3). For the PPO agent, the state space is defined as $\{B_y, P_z, S_x, a_h\}$, where $a_h$ are the actions of the human. The action space is one of the available actions (i.e., translation or rotation). The hyperparameters of the PPO agent are shown in Table I. Each training episode was 40 seconds to avoid decreased human performance due to the human's variation or frustration.

Three learning strategies are validated in the designed experiment, include the two proposed learning strategies and a baseline strategy. Each learning strategy has two training cases by swapping the robot's and human's actions. In total, 6 training cases are designed. They are responsibility-guided cooperative learning (learn the task first: case 1 and case 2), level-guided cooperative learning (learn the human first: case 3 and case 4) and learn all tasks together (Baseline: case 5, and case 6). In cases 1, 3, and 5, the human controls the rotation action, and the robot controls the translation action. In cases 2, 4, and 6, the human and the robot swap actions. 6 human subjects were invited to the experiments. Each human subject only completed one training case and was not involved in other cases to ensure that the human subjects had no previous knowledge of the experiment setup for proper validation. The details of reward structures and training steps for cases 1, 3, and 5 are outlined next. The task decomposition trees of cases 1, 3, and 5 are shown in Fig. 4. For simplicity, the details of cases 2, 4, and 6 are not shown because they follow the same structures as cases 1, 3, 5.

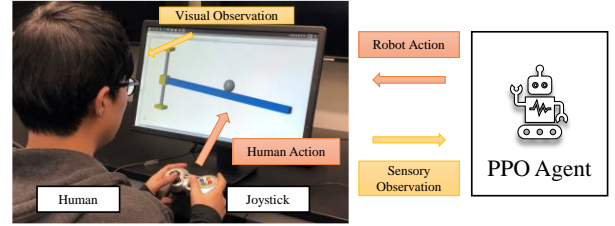

Fig. 3. Experiment setup. The robot gets sensory observations, including human action and state information. The human can observe the real-time simulated results shown on the monitor. The robot directly inputs its action to the model, and the human input his/her action to the model by controlling a joystick.

TABLE I. HYPER-PARAMETERS FOR PPO ALGORITHM

| Parameters | Values |
| --- | --- |
| Discount Factor($\gamma$) | 0.995 |
| Experience Horizon | 512 |
| Entropy Loss Weight | 0.02 |
| Clip Factor | 0.05 |
| GAE Factor | 0.95 |
| Sample Time | 0.2 |
| Mini-Batch Size | 64 |
| Learning Rate | 0.001 |
| Number of Epoch | 3 |

**Case 1:** In the first training step, the robot learns the slider position subtask and pendulum position subtask without human involvement. The reward function is:

$$R_1|case\ 1 = f_1^2 + f_2^2 \tag{6}$$

In the second training step, the human joins the training. The robot and the human learn the overarching task and cooperate. The robot follows the reward:

$$R_2|case\ 1 = f_1^2 + f_2^2 + f_1^1 \tag{7}$$

**Case 3:** In the first training step, the robot and the human are trained together to learn the lower-level subtasks, where the robot learns the slider task with reward function:

$$R_1|case\ 3 = f_1^2 + f_2^2 \tag{8}$$

The human learns the pendulum task. In the second training step, the team cooperatively learns all subtasks. The robot's reward function becomes:

$$R_2|case\ 3 = f_1^2 + f_2^2 + f_1^1 \tag{9}$$

**Case 5:** The robot and the human train together to learn the overarching task, where the robot's reward function is:

$$R|case\ 3 = f_1^2 + f_2^2 + f_1^1 \tag{10}$$

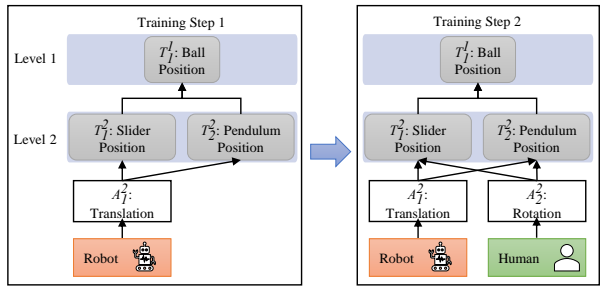

(a) Responsibility-guided cooperative learning (learn the task first).

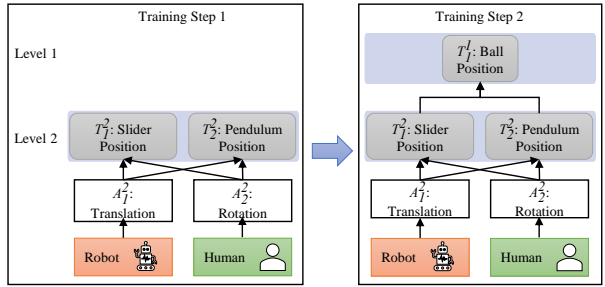

(b) Level-guided cooperative learning (Learn the human first).

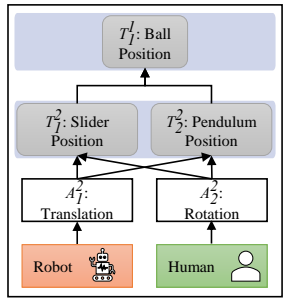

(c) Learn together.

Fig. 4. (a), (b) and (c) are the task decomposition trees of each training step for cases 1, 3, and 5. Cases 2, 4, and 6 have similar structures with swapped tasks between the human and the robot.

## V. RESULTS AND DISCUSSION

### A. Training Processes and Performance Evaluation

Fig. 5 shows the results of the training process for all cases. For the two proposed learning strategies, both case 1 and case 4 converged in less than 7 iterations, which are faster than the other cases. Case 2 struggled the most (when the robot learned the task alone). In case 3, the robot met the difficulty of learning initially but still reached the target 40% faster than case 2. After the second training step, case 4 converged in 14 iterations, which is the fastest. Case 2 converged in 31 iterations, which is the slowest. In the baseline strategy, case 6 successfully reached the convergence condition in 16 iterations, but case 5 failed the training because the team could not reach the convergence condition, and the team performance started to decrease. The robot's action and the human's action in 4 seconds during a trial of cooperation after the training are shown in Fig. 6 to analyze the team performance and better understand the robot's behaviors and the human.

Table II shows the training processes' statistics, the performance evaluation of the trained HRC policies and the effort of the human and the robot. The training process section includes the number of human-involved iterations, total iterations, and corresponding proportions. The performance evaluation includes the cumulative numerical error for the low- and high-level subtasks and total error. The effort is evaluated by calculating the variance of the actions. High variance means it takes more effort for human/robot to adapt to the partner and maintain the task performance. Case 2 is highlighted as the best performing because it had the least human-involved training in terms of the number of training iterations and proportion to total training iterations. It also achieved the least errors in both the low-level slider task and the high-level ball balance task. Case 2 still achieved the second-best performance for the low-level pendulum task with only an 8% performance gap with case 4. Case 2 also has the least human effort, and the robot efforts are similar across all case due to the consistency of the learning approach. In case 1, the human had to be more involved to reach a mediocre performance. Cases 3-6 needed 100% human involvement. Case 4 achieved the second-best performance, and case 5 had the worst performance.

### B. Influence of Human and Asymmetric Hierarchical Task

The results show that the proposed method helped the robot successfully learn the HRC task in cases 1 - 4 and 6. The failure of case 5 is due to the inherent low learning efficiency problems that occurred when the robot must learn the difficult task and the human together. In cases 2, 4, and 6, where the human controls the translation action and the robot control the rotation action, the average total error was 64% less than in cases 1, 3, and 5, where the roles were swapped. Fig. 6 also shows that in cases 1, 3, and 5, the robot must adjust its action to accommodate human action changes. Such adjustment costs extra effort for both the human and robot, downgrading learning efficiency and team performance. The harmonious actions of both the human and the robot in cases 2, 4, and 6 indicate that the robot has found a good cooperative policy and can reach a higher performance.

A practical reason for the poor team performance of cases 1, 3, and 5 is that the human perception is better in observing translational movement than rotational movement. In cases 1, 3, and 5, the human struggled to identify if the pendulum was in the horizontal direction and therefore took a lot of effort adjusting the pendulum position. The human adjustment then affected the robot's observations and forced the robot to accommodate. In cases 2, 4, and 6, the human can adequately estimate the slider's translational movement and apply appropriate action. The comparison shows the assignment of the human actions and tasks will end up with different team performance. With this understanding, we believe task and role allocation is another promising topic for HRC.

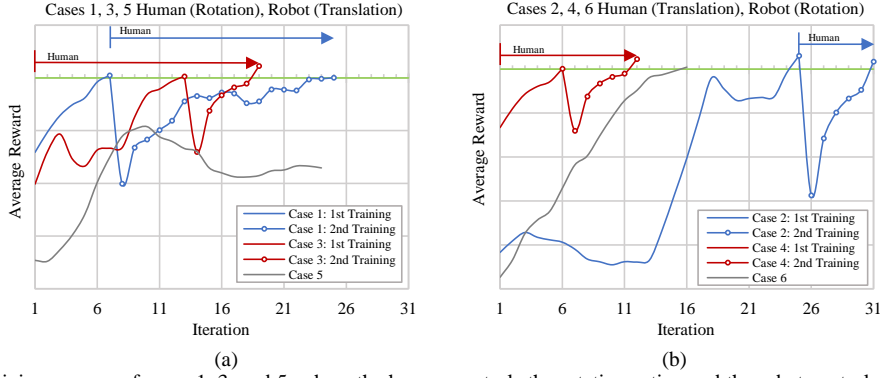

Fig 5. (a) shows the training process of cases 1, 3, and 5, where the human controls the rotation action and the robot controls the translation action. (b) shows the training process core cases 2, 4, and 6 where the actions are swapped. For Responsibility-guided and Level-guided strategies, the reward functions are updated while finishing the first training step, then starts the second training step, which results in a reward drop. The times where the human joins the training are marked with the red and blue arrows.

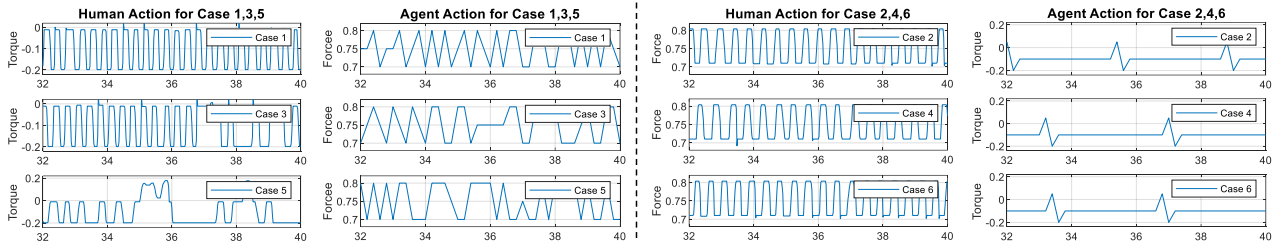

Fig 6. The actions were executed by the human and the robot during 4 seconds of the policy validation process. In cases 1, 3, and 5, the human controls the rotation action, and the robot controls the translation action. In cases 2, 4, and 6, the actions are swapped. For cases 1, 3, and 5, only case 1 can learn a relatively good policy with clearer action patterns for the human and robot. In case 3 and case 5, the human and the robot struggled and must adjust their actions. On the contrary, in cases 2, 4, and 6, the human and robot learned harmonious action patterns with much lower action frequency.

TABLE II. TRAINING PROCESS (EPISODE), MODEL PERFORMANCE (ERROR) AND EFFORT

| Category & Case | | Training (Episode) | | | Performance (Error) | | | | Effort | |
|---|---|---|---|---|---|---|---|---|---|---|
| | | Human Involved | Total | Percentage | Slider (m) | Pendulum (rad) | Ball (m) | Total | Human | Robot |
| Learn Task First | Case 1 | 18 | 25 | 72% | 0.26 | 0.57 | 0.78 | 1.61 | 0.0075 | 0.0022 |
| | Case 2 | 6 | 31 | 19% | 0.07 | 0.38 | 0.37 | 0.82 | 0.0020 | 0.0023 |
| Learn Human First | Case 3 | 19 | 19 | 100% | 0.17 | 0.85 | 1.01 | 2.04 | 0.0078 | 0.0016 |
| | Case 4 | 14 | 14 | 100% | 0.11 | 0.35 | 0.69 | 1.14 | 0.0021 | 0.0022 |
| Learn Together | Case 5 | 24 | 24 | 100% | 0.28 | 3.16 | 2.82 | 6.26 | 0.0096 | 0.0019 |
| | Case 6 | 16 | 16 | 100% | 0.13 | 0.41 | 0.70 | 1.24 | 0.0023 | 0.0024 |

### C. Learning Strategies for Cooperative Robot in HRC

The training processes in Fig. 5 and the statistical results in Table II confirm our hypothesis that **robots should separately learn the task and human partner.** Regarding the question of what the robot should learn first, we realized that there are multiple answers.

Specifically, in the responsibility-guided learning strategy (learn task first), the human's non-involvement helps the robot better learn the task's dynamics. The robot can fully observe the environment in the first training step without human influence, making it easier for the robot to explore the environment. When the human joins the task, the robot only needs to learn human behavior and update its policy to cooperate. The level-guided learning strategy (learn human first) achieves higher learning efficiency, mainly because human involvement helps the robot narrow down the exploration space. However, the human sacrifice his/her effort to help the robot learn as the human involvement increases. The difficulty still exists for the robot to decompose the end effects of human action and robot action in the high-level task. That is a possible reason that this strategy achieved lower team performance.

The baseline strategy (learn together) proves our analysis on the drawbacks of the state-of-art methods where the training would be affected by the human's variation and the robot's random exploration. Case 6 managed to converge because the actions were appropriately assigned to the human and the robot. Case 5 shows that the tedious robot exploration causes frustration to the human, and the human's mistakes misled the robot during the learning process. Consequently, team performance is hard to increase.

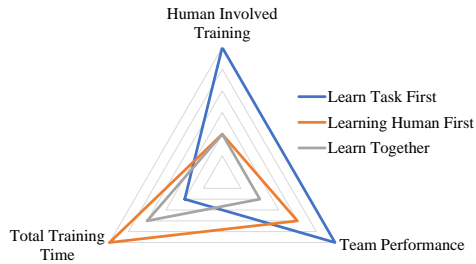

Fig 7. A three-factor diagram considers human involvement, total training time, and team performance for the validated training categories. The outer side means higher team performance, less human-involved training, and less total training time.

Conventional performance evaluation metrics only consider team performance and total training time. However, we believe that another important factor needs to be considered, which is human-involved training. Ideally, we want to reduce the burden on the human in HRC. Comparing the three learning strategies is shown in Fig. 7 as a three-factor diagram, which is convenient to select the desired learning strategy. For example, if faster training is the priority, the robot should learn the human first. If good team performance is the priority, the robot should learn the task first. If minimal human involvement is the priority, the robot should learn the task first.

## VI. CONCLUSION

In this work, with the proposed task decomposition method and hierarchical reward mechanism, we studied the nature of HRC in a hierarchical dynamic control task with a hierarchical structure. Experiment results demonstrate that robots should learn the task and human partners separately. A three-factor performance evaluation metrics were introduced to achieve careful consideration for learning strategy selection. Our future work will focus on the task and role allocation for a better cooperative robot with adaptability, self-awareness, and partner-awareness.

## ACKNOWLEDGMENT

This material is based on work supported by the US NSF under grant 1652454 and 2114464. Any opinions, findings, conclusions, or recommendations expressed in this material are those of the authors and do not necessarily reflect those of the National Science Foundation.